%% file: main.tex
\documentclass[10pt,twocolumn,letterpaper]{article}

\usepackage[accsupp]{axessibility}
\usepackage[pagenumbers]{iccv} 


\definecolor{iccvblue}{rgb}{0.21,0.49,0.74}
\usepackage[pagebackref,breaklinks,colorlinks,allcolors=iccvblue]{hyperref}

\usepackage{multirow}

\newcommand{\bblue}[1]{#1}

\def\paperID{13298} 
\def\confName{ICCV}
\def\confYear{2025}

\title{Correspondence as Video: Test-Time Adaption on SAM2\\for Reference Segmentation in the Wild}

\author{
    Haoran Wang$^{1}$ \quad
    Zekun Li$^{1}$ \quad 
    Jian Zhang$^{1}$ \quad
    Lei Qi$^{2}$ \quad
    Yinghuan Shi$^{1}$\thanks{Corresponding author: Yinghuan Shi. This work is supported by NSFC Project (62192783, 62206052, 62222604), China Postdoctoral Science Foundation (2024M750424), Fundamental Research Funds for the Central Universities (020214380120, 020214380128), State Key Laboratory Fund (ZZKT2024A14, ZZKT2025B05), Postdoctoral Fellowship Program of CPSF (GZC20240252), Jiangsu Funding Program for Excellent Postdoctoral Talent (2024ZB242) and Jiangsu Science and Technology Major Project (BG2024031).} \quad
    \\
  $^{1}$Nanjing University \quad $^{2}$Southeast University \\ 
}

\begin{document}
\maketitle

\input{sec/0_abstract}

\input{sec/1_intro}

\input{sec/2_related_works}

\input{sec/3_method}

\input{sec/4_experiments}

\input{sec/5_conclusion}

\clearpage
{
    \small
    \bibliographystyle{ieeenat_fullname}
    \bibliography{main}
}

\end{document}

%% file: sec/0_abstract.tex
\begin{abstract}
Large vision models like the Segment Anything Model (SAM) exhibit significant limitations when applied to downstream tasks in the wild. Consequently, reference segmentation, which leverages reference images and their corresponding masks to impart novel knowledge to the model, emerges as a promising new direction for adapting large vision models. However, existing reference segmentation approaches predominantly rely on meta-learning, which still necessitates an extensive meta-training process and brings massive data and computational cost.
In this study, we propose a novel approach by representing the inherent correspondence between reference-target image pairs as a pseudo video. This perspective allows the latest version of SAM, known as SAM2, which is equipped with interactive video object segmentation (iVOS) capabilities, to be adapted to downstream tasks in a lightweight manner. We term this approach \textbf{C}orrespondence \textbf{A}s \textbf{V}ideo for \textbf{SAM} (CAV-SAM). CAV-SAM comprises two key modules: the \textbf{D}iffusion-\textbf{B}ased \textbf{S}emantic \textbf{T}ransition (\textbf{DBST}) module employs a diffusion model to construct a semantic transformation sequence, while the \textbf{T}est-\textbf{T}ime \textbf{G}eometric \textbf{A}lignment (\textbf{TTGA}) module aligns the geometric changes within this sequence through test-time fine-tuning.
We evaluated CAV-SAM on widely-used datasets, achieving segmentation performance improvements exceeding 5\% over SOTA methods.
Our implementation is available at \url{https://github.com/wanghr64/cav-sam}.
\end{abstract}

%% file: sec/1_intro.tex
\section{Introduction}

\begin{figure}[t]
    \centering
    \includegraphics[width=\linewidth]{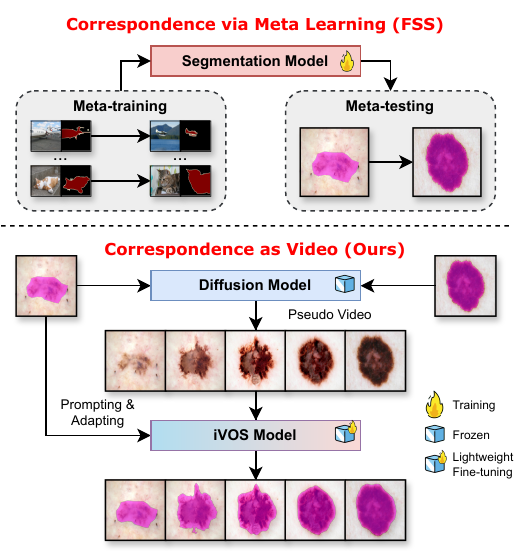}
    \caption{\textbf{Different approaches for reference segmentation in the wild.} Traditional methods like FSS rely on extensive meta-training, incurring significant data and computational burdens. In contrast, we embody the correspondence between images as a generated video and perform segmentation using an iVOS model with only lightweight fine-tuning throughout the process.}
    \label{fig:new-intro-1}
\end{figure}
 
Recent large vision models, \eg, the Segment Anything Model (SAM)~\cite{kirillov2023segment}, exhibit impressive segmentation performance and generalization abilities. However, for various segmentation tasks in the wild, the domain gap and class novelty between downstream datasets and SAM's training datasets frequently result in decreased segmentation accuracy~\cite{zhang2023comprehensive,mazurowski2023segment}.
To overcome these challenges, reference segmentation emerges as a promising approach. 
Reference segmentation employs a small set of annotated images from downstream tasks as the \emph{reference images}, enabling the model to understand the novel domain and class characteristics, facilitating the segmentation of the \emph{target images} accordingly.
\bblue{Few-shot segmentation (FSS)~\cite{lang2023base,liu2022intermediate,cheng2022holistic,peng2023hierarchical,wang2023rethinking} and cross-domain few-shot segmentation (CD-FSS)~\cite{lei2022cross,wang2022remember,herzog2024adapt,he2024apseg,su2024domain,nie2024cross} represent specific sub-settings within reference segmentation.}
Specifically, FSS focuses on learning new classes, while CD-FSS enhances robustness to domain variations. Additionally, some approaches~\cite{he2024apseg,sun2024vrp} have applied reference segmentation directly to SAM. 
However, as illustrated in the upper part of Fig.~\ref{fig:new-intro-1}, these approaches predominantly rely on extensive meta-training to acquire class-agnostic and domain-agnostic knowledge, necessitating considerable extra computational resources.

\begin{figure}[t]
    \centering
    \includegraphics[width=1\linewidth]{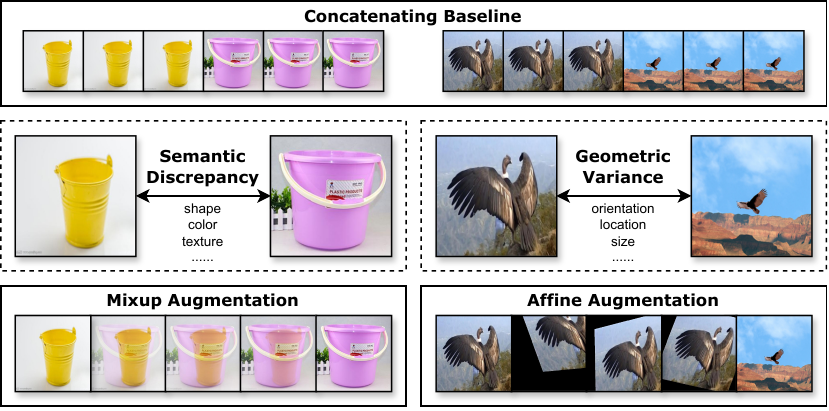}
    \caption{\textbf{Challenges of directly utilizing iVOS model for reference segmentation and their heuristic solutions.} Discrete image pairs often display substantial semantic discrepancies and geometric variations. Two heuristic solutions proposed to address these challenges respectively---affine and mixup augmentation---fail to adequately simulate the natural transition of videos, rendering them ineffective in resolving these challenges.}
    \label{fig:challenges}
\end{figure}

To avoid the substantial overhead associated with meta-training, we propose a new perspective for reference segmentation.
We observe that there is existing \emph{correspondence} between reference and target images, which is not yet fully leveraged due to the discrete nature of reference-target images. 
Fortunately, recent advancements in \textbf{i}nteractive \textbf{V}ideo \textbf{O}bject \textbf{S}egmentation (\textbf{iVOS}) models~\cite{cheng2023tracking,yang2023track,cheng2023segment,rajivc2023segment}, \eg, the latest version of SAM, SAM2~\cite{ravi2024sam}, make it possible to directly capture the correspondence between images. 
These iVOS models segment an entire video sequence by annotating certain frames. 
We experimented that as shown in Fig.~\ref{fig:challenges}, by simply concatenating reference-target images to construct a pseudo video sequence, SAM2 achieves near the SOTA performance (about 61) with mIoU 60.68 on 1-shot CD-FSS benchmarks, demonstrating the significant potential of the iVOS model when applied to reference segmentation tasks. 
This highlights the significant potential of iVOS models for reference segmentation tasks in the wild.

However, this concatenating baseline method still offers considerable potential for enhancement. This is primarily because discrete image pairs, unlike natural continuous video sequences, exhibit the following characteristics as depicted in Fig.~\ref{fig:challenges}:
\textbf{a) Semantic Discrepancy:} iVOS model's video segmentation is specifically designed to track the same object instance across a video sequence, where the semantic of the object are largely consistent over time. 
In contrast, reference segmentation focuses on identifying objects belonging to the same class rather than the same instance. 
This fundamental difference results in significant intra-class semantic discrepancy among target objects in a reference-target pair, thereby complicating the effective application of iVOS model's video segmentation techniques to reference segmentation.
\textbf{b) Geometric Variation:}
The effectiveness of iVOS model in video segmentation is largely due to its optimization for objects that exhibit smooth geometric transformations across frames. 
In reference segmentation, however, the reference and target images may present the target object with considerable geometric variation. 
Such variations hinder iVOS model's capability to perceive these variations as the same object transitioning through a video sequence, thereby challenging its adaptability to various downstream reference segmentation task.

The key challenge, therefore, lies in how to transform discrete reference-target image pairs into a smooth pseudo video sequence. We experimented with two simple heuristic methods, mixup and affine augmentation to address the two challenges respectively. But these heuristic approaches failed and even led to a decline in performance compared to the concatenation baseline. We believe this is because the pseudo video sequence generated by heuristic methods is still far from natural.
To address this, we instead treat the \textbf{C}orrespondence between reference-target images \textbf{A}s \textbf{V}ideo, introducing \textbf{CAV-SAM} with two modules to explore more fundamental solutions:
a) \textbf{D}iffusion-\textbf{B}ased \textbf{S}emantic \textbf{T}ransition (\textbf{DBST}): Recent advances in diffusion-based generative models have shown great promise in enhancing data diversity. Built on recent DiffMorpher~\cite{zhang2024diffmorpher}, we utilize diffusion model to generate a series of semantically meaningful transition sequences between reference-target image pairs, facilitating SAM2's ability to capture smoother semantic changes in reference-target objects.
b) \textbf{T}est-\textbf{T}ime \textbf{G}eometric \textbf{A}lignment (\textbf{TTGA}): Instead of extensive meta-training, we perform lightweight test-time adaptation of SAM2 using only the one reference image. We activate the target objects within the sequence using prototype vectors. The activations will serve as additional prompts for SAM2 to better align with the geometric variance of the target objects. Our main contributions are summarized as follows:
\begin{itemize}
    \item Our innovative approach to reference segmentation shifts away from the meta-learning framework by treating the discrete correspondence between reference-target images as a continuous pseudo video sequence, allowing iVOS models to effortlessly adapt to various downstream tasks.
    \item We utilize diffusion model to generate a smooth semantic transition sequence between reference-target images. Additionally, we apply lightweight test-time adaptation to make the iVOS model align with drastic geometric changes in target objects within the transition sequence.
    \item We conducted extensive evaluations on four datasets of the popular CD-FSS benchmark. Our approach achieved approximately 5\% improvement in mIoU over existing SOTA methods, demonstrating exceptional performance on challenging in-the-wild datasets such as Chest X-Ray.
\end{itemize}

%% file: sec/2_related_works.tex
\section{Related Work}

\begin{figure*}[t]
    \centering
    \includegraphics[width=\linewidth]{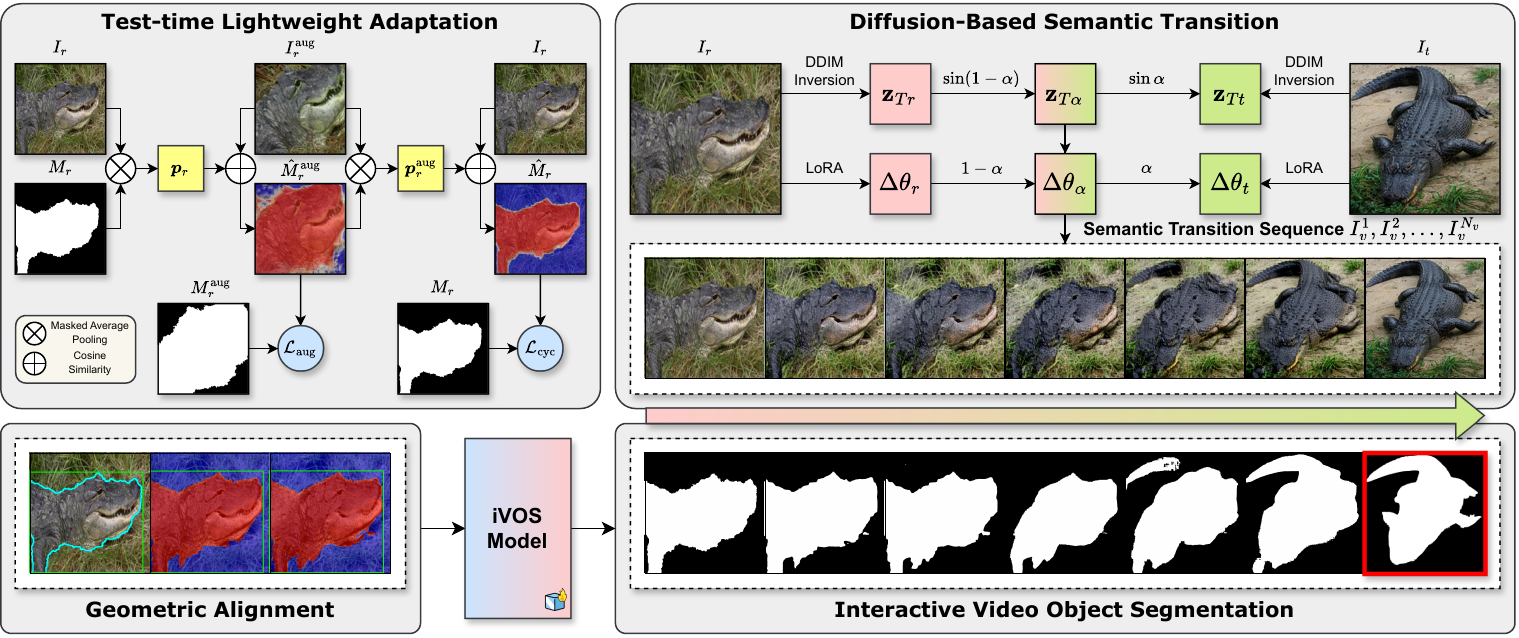}
    \caption{\textbf{Overall architecture of the proposed CAV-SAM model.} The \textbf{DBST} (\textbf{D}iffusion-\textbf{B}ased \textbf{S}emantic \textbf{T}ransition) module uses diffusion model to interpolate semantics between reference and target images, allowing smoother transitions in generated sequences that resemble video-like footage. The \textbf{TTGA} (\textbf{T}est-\textbf{T}ime \textbf{G}eometric \textbf{A}lignment) module employs prototype vectors and augmented images for lightweight fine-tuning, to obtain extra prompts for SAM2 to align with geometric variations within the pseudo video sequence.}
    \label{fig:method-overview}
\end{figure*}

\subsection{Reference Segmentation}
\noindent\textbf{Few-shot Segmentation}
aims to segment novel categories with minimal labeled examples through dense mask predictions. Methods are primarily divided into prototype-based and matching-based approaches. 
Prototype-based methods~\cite{lang2022learning,lang2023base,li2021adaptive,liu2022intermediate,tian2020prior,cheng2022holistic} use class prototypes from support images that represent target objects.
Matching-based methods~\cite{peng2023hierarchical,wang2023rethinking,lu2021simpler,min2021hypercorrelation,zhang2021few} focus on dense correspondences between query and support features, enhancing pixel-level detail and contextual understanding. 
However, all these approaches face challenges in cross-domain generalization. 

\noindent\textbf{Cross-Domain Few-Shot Segmentation}
~\cite{lei2022cross,wang2022remember,herzog2024adapt,he2024apseg,su2024domain,nie2024cross} has gained attention due to the challenge of training models without access to target data.
Recent approach like IFA~\cite{nie2024cross} introduces a strategy through bi-directional prediction and iterative adaptor, and DR-Adaptor~\cite{su2024domain} leverages a domain-rectifying adapter featuring local-global style perturbation and cyclic domain alignment mechanism.

\noindent\textbf{SAM for Reference Segmentation.}
SAM~\cite{kirillov2023segment} was designed for interactive segmentation and does not natively support reference segmentation. Recent approaches, \eg, APSeg~\cite{he2024apseg} and VRP-SAM~\cite{sun2024vrp} have incorporated meta-learning techniques by introducing an additional handcrafted prompt encoder module, which generates meta-prompts for SAM based on the reference images.

\noindent\textbf{Comparison with Our Approach.}
Most of the reference segmentation methods mentioned above follow the meta-learning paradigm, requiring extensive episodic meta-training on massive datasets. In contrast, our method fully utilizes the inherent correspondence between reference-target images. By applying test-time lightweight fine-tuning using only one reference image, we significantly reduce both data and computational requirements.

\subsection{Interactive Video Object Segmentation.}
\noindent\textbf{Segment Anything Model 2}
~\cite{ravi2024sam} builds upon the original Segment Anything Model~\cite{kirillov2023segment} to address the challenges posed by videos.
SAM2 enhances segmentation capabilities by incorporating a memory-augmented architecture.
Using prompts such as points, boxes, or masks on any video frame, SAM2 predicts a spatio-temporal mask sequence.

\noindent\textbf{Connection to Our Approach.}
iVOS models excel at capturing relationships between video frames, making them well-suited for our approach to reference segmentation. By converting a discrete image pair into a semantically continuous pseudo video sequence using diffusion model, iVOS models can deliver excellent segmentation performance.

%% file: sec/3_method.tex
\section{Method}

\subsection{Problem Formulation}

In reference segmentation, the reference image \( I_r \) and its mask \( M_r \) are used to segment the target image \( I_t \). As illustrated in Fig.~\ref{fig:method-overview}, our approach interprets the correspondence between the reference-target images as a video, by representing it as a semantic transformation sequence \( I_v^1, I_v^2, \ldots, I_v^{N_v} \) using diffusion model. During test-time, we perform a lightweight adaptation using only the reference image \( I_r \) and its mask \( M_r \), along with the augmented image \( I_r^{\text{aug}} \) and its augmentative pixel-corresponding mask \( M_r^{\text{aug}} \). After adaptation, the model utilizes the prototype vector \( p_r \) obtained from \(I_r\) and \(M_r\) to activate frames \( I_v^i \) within the sequence and generate pseudo-labels, which serve as extra prompts for the SAM2 model, facilitating further geometric alignment of pseudo video sequence.

\subsection{Diffusion-Based Semantic Transition}

\begin{figure}
    \centering
    \includegraphics[width=1\linewidth]{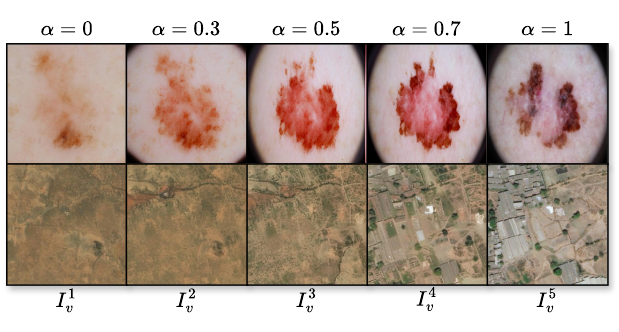}
    \caption{\textbf{Semantic transition sequence generated by the DBST module on ISIC and Deepglobe dataset}. By adjusting the interpolation ratio $\alpha$, the DBST module produces intermediate images $I_v^1,I_v^2,\dots,I_v^{N_v}$ that demonstrate varying semantic elements between the reference-target images $I_r$ and $I_t$.}
    \label{fig:morpher-examples}
\end{figure}

SAM2's video segmentation focuses on semantically consistent object instances, whereas FSS tasks require identifying objects of the same class, presenting challenges due to intra-class semantic discrepancies.
Built on DiffMorpher~\cite{zhang2024diffmorpher}, our DBST module interpolates LoRA parameter \(\Delta \theta\) and latent noise \(\mathbf{z}_{Tt}\) between reference-target images by the ratio \(\alpha\), generating pseudo video frames $I_v^1,I_v^2,\dots,I_v^{N_v}$ illustrating semantic transitions.
By eliminating the unnecessary refinement modules in DiffMorpher, we significantly accelerate the generation process.
As shown in Fig.~\ref{fig:morpher-examples}, by adjusting different $\alpha$ values, our DBST module can generate frames with smooth semantic transition between images.

\textbf{Preliminary: the Diffusion Model.}
The diffusion model~\cite{ho2020denoising,song2020denoising,song2020score} is a generative framework with two main stages: a forward process that adds noise incrementally to a clean sample, and a backward process that reconstructs the sample.
The forward process \(\{ q(\mathbf{z}_t) \mid t=0,1,\dots,T \}\) gradually adds noise to data from distribution \(q(\mathbf{z}_0)\), resulting in \(q(\mathbf{z}_T) = \mathcal{N}(\mathbf{0}, \mathbf{I})\).
The backward process \(\{ p(\mathbf{z}_t) \mid t=T, T-1, \dots, 0 \}\) reconstructs the sample from Gaussian noise, with \(\mathbf{z}_T \sim \mathcal{N}(\mathbf{0}, \mathbf{I})\).
The denoising is parameterized by a noise prediction network \(\varepsilon_\theta\), which predicts the noise \(\varepsilon\) added to \(\mathbf{z}_{t-1}\) using the current noise \(\mathbf{z}_t\) and time step \(t\).

\textbf{LoRA Parameter Interpolation.}
LoRA~\cite{hu2021lora} efficiently tunes diffusion models by adjusting parameters \(\theta\) via a low-rank residual \(\Delta \theta\).
This approach captures image semantics in the parameter space.
By fitting LoRA to the reference-target images \((I_r,I_t)\), yielding \((\Delta \theta_r,\Delta \theta_t)\), semantics of these images are fused through linear interpolation:
\begin{equation}
\Delta \theta_\alpha = (1-\alpha)\Delta \theta_r + \alpha \Delta \theta_t.
\end{equation}
The interpolated \(\Delta \theta_\alpha\) is used in the noise prediction network during the denoising process.

\textbf{Latent Noise Interpolation.}
For the latent noises of the reference and target images \(\mathbf{z}_{Tr}\) and \(\mathbf{z}_{Tt}\) obtained from DDIM~\cite{song2020denoising} inversion, we derive an intermediate latent noise \(\mathbf{z}_{T\alpha}\) using spherical linear interpolation:
\begin{equation}
\mathbf{z}_{T\alpha} = \frac{\sin((1-\alpha)\phi)}{\sin \phi}\mathbf{z}_{Tr} + \frac{\sin(\alpha\phi)}{\sin\phi}\mathbf{z}_{Tt}.
\end{equation}
The latent noise \(\mathbf{z}_{T\alpha}\) is then denoised using the LoRA-integrated noise prediction network parameterized by \(\epsilon_{\theta+\Delta\theta_\alpha}\) with a DDIM schedule, resulting in intermediate images $I_v^1,I_v^2,\dots,I_v^{N_v}$ with natural semantic transitions.

\textbf{Optimizations over vanilla Diffmorpher.}
The primary goal of DiffMorpher~\cite{zhang2024diffmorpher} is to generate intermediate transition images that are more \emph{visually meaningful to humans}, which leads to various extra refinement modules.
However, basic low-level semantic transitions are already sufficient for SAM2 thanks to its robust iVOS capabilities. 
The pursuit of excessive visual detail would only result in an intolerable computational burden. 
Consequently, we removed these unnecessary modules and optimized the parameters for LoRA training and DDIM inversion, significantly reducing the inference costs of the DBST module.
In Section~\ref{sec:parameter-sensitivity}, we demonstrate that while the outputs generated by our lightweight DBST module may be less visually appealing compared to the vanilla DiffMorpher, the segmentation performance remains consistently excellent.

\subsection{Test-time Geometric Reinforcement}

While our DBST module bridges the semantic discrepancy, excessive and unnatural geometric variances within reference-target images can further hinder the instance-tracking efficacy of SAM2. 
Our TTGA module refines class prototype vector $\boldsymbol{p}_r$ at test-time using only the reference image \( I_r \) and its mask \( M_r \), along with the augmented image \( I_r^{\text{aug}} \) and its augmentative pixel-corresponding mask \( M_r^{\text{aug}} \). 
The refined prototype vectors activate the semantic transformation sequence $I_v^1,I_v^2,\dots,I_v^{N_v}$, serving as extra prompts for SAM2 to further align the geometric variations.

\textbf{Preliminary: Prototype Vector.}
The model uses the reference image feature \( F_r\in\mathbb R^{ H\times W\times D} \) and its mask \( M_r\in\mathbb R^{H\times W} \) to derive a representative prototype vector \( \boldsymbol{p}_r \in\mathbb R^{D} \) via masked average pooling (MAP):
\begin{equation}
\label{eq:proto}
    \boldsymbol{p}_r =\mathrm{MAP}(F_r,M_r)= \frac{\sum_{i}^{H} \sum_{j}^{W} F_r[ i, j,:] \cdot M_r[i, j]}{\sum_{i}^{H} \sum_{j}^{W} M_r[i, j]}.
\end{equation}
The cosine similarity between the prototype $\boldsymbol{p}_r$ and target feature \( F_t\) is then used to segment the target image:
\begin{equation}
\label{eq:similarity}
    S_t = \frac{F_t \cdot \boldsymbol{p}_r}{\| F_t \|_2 \| \boldsymbol{p}_r \|_2}.
\end{equation}
Since the resulting similarity map \( S_t \in\mathbb R^{H\times W}\) is not probabilistic, the threshold is determined by~\cite{otsu1975threshold}, producing the final target image segmentation result \( \hat{M}_t \in\mathbb R^{H\times W}\):
\begin{equation}
\label{eq:otsu}
    \hat{M}_t = \mathbb{I}(S_t > \tau), \quad \tau = \mathrm{otsu}(S_t),
\end{equation}
where $\mathbb{I}$ denotes the indicator function, \bblue{and $\mathrm{otsu}(\cdot)$ is the thresholding algorithm for grayscale image segmentation.}

\begin{figure}[t]
    \centering
    \includegraphics[width=1\linewidth]{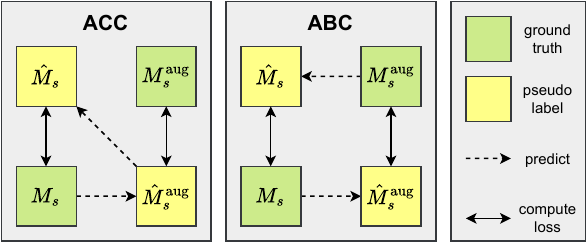}
    \caption{\textbf{Comparative diagram illustrating ACC and ABC.} Both methods utilize original reference image $I_r$ with its mask $M_r$ to generate pseudo-label $\hat M_r^{\text{aug}}$ for the augmented $I^{\text{aug}}_r$. ACC employs the pseudo-label $\hat M_r^{\text{aug}}$ while ABC uses the mask $M_r^{\text{aug}}$.}
    \label{fig:acc-abc}
\end{figure}

\textbf{Test-time Fine-tuning.}
Direct extraction of image features $F_r$ from SAM2 image encoder inadequately activates the semantic transition sequence $I_v^1,I_v^2,\dots,I_v^{N_v}$, making the fine-tuning essential. 
We streamlined the fine-tuning process from \emph{architecture} and \emph{data} perspectives.

From the architectural perspective, we exclusively fine-tune \bblue{the FPN layer} \verb|neck| of the SAM2 image encoder.
From the data perspective, we utilize only the reference image $I_r$ and its corresponding mask $M_r$ from the FSS episode.
To maximize the utility of the limited labeled data, we introduce \emph{augmented} reference image $I_r^{\text{aug}}$ and its corresponding mask $M_r^{\text{aug}}$ with \textbf{A}ugmentative \textbf{C}yclic \textbf{C}onsistency (\textbf{ACC}) as the learning objective as shown in ~\ref{fig:acc-abc}. 
Specifically, we first derive the prototype vector $\boldsymbol p_r$ using $I_r$ and $M_r$ as per Eq.~\ref{eq:proto}. 
Then we obtain the predicted mask $\hat M_r^{\text{aug}}$ with $\boldsymbol p_r$ according to Eq.~\ref{eq:similarity} and \ref{eq:otsu}. Given that the augmented reference image $I_r^{\text{aug}}$ has its corresponding mask $M_r^{\text{aug}}$ as the ground truth, we compute the loss as follows:
\begin{equation}
    \mathcal{L}_{\text{aug}} = \mathrm{BCE}(\mathrm{sigmoid}(S_r^{\text{aug}}), M_r^{\text{aug}}),
\end{equation}
where $\mathrm{BCE}(\cdot)$ denotes binary cross entropy.

\begin{figure}[t]
    \centering
    \includegraphics[width=\linewidth]{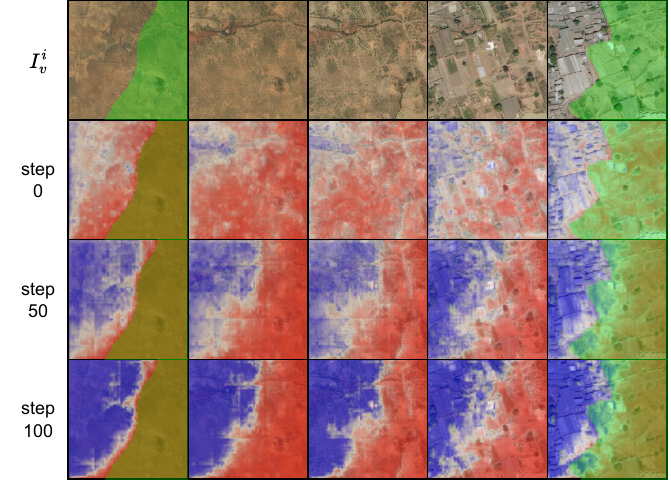}
    \caption{\textbf{Illustration of our lightweight test-time fine-tuning process.} Prototype vector $\boldsymbol p_r$ is iteratively refined for activations on $I_v^1,I_v^2,\dots,I_v^{N_v}$, which will serve as extra prompts for SAM2 to align with the geometric variance within the frames.}
    \label{fig:TTGA-fine-tune}
\end{figure}

\begin{figure}[t]
    \centering
    \includegraphics[width=\linewidth]{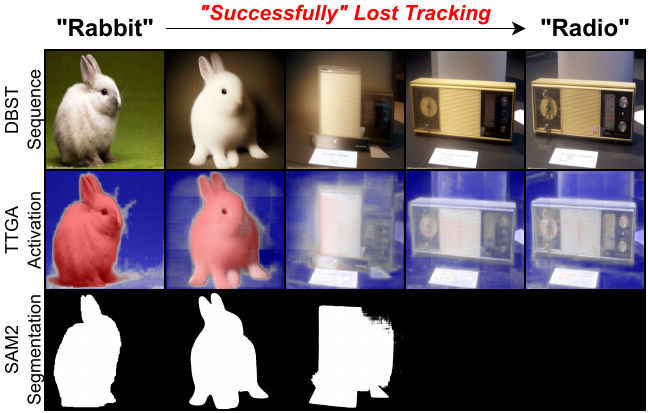}
    \caption{Illustration of our proposed method ensuring semantic consistency. In cases where the classes of reference and target images differ, both DBST and TTGA modules guarantee the iVOS model achieves the wanted ``failure''---producing no segmentation.}
    \label{fig:semantic_consistency}
\end{figure}

With $I_r^{\text{aug}}$ and its predicted mask $\hat M_r^{\text{aug}}$, we obtain the augmented prototype vector $\hat{\boldsymbol p}_r^{\text{aug}}$, and predict $\hat M_r$ on the original $I_r$. We define the cyclic loss as follows:
\begin{equation}
    \mathcal{L}_{\text{cyc}} = \mathrm{BCE}(\mathrm{sigmoid}(S_r), M_r).
\end{equation}
The final loss for is defined as the sum of these two losses:
\begin{equation}
    \mathcal{L} = \mathcal{L}_{\text{aug}} + \mathcal{L}_{\text{cyc}}.
\end{equation}

Additionally, we also investigate an alternative strategy \textbf{A}ugmentative \textbf{B}i-directional \textbf{C}onsistency (\textbf{ABC}) as depicted in Fig.~\ref{fig:acc-abc}. Unlike ACC, ABC utilizes the ground truth mask \(M_r^{\text{aug}}\) of the augmented image to compute \(\boldsymbol p_r^{\text{aug}}\), while the remaining processes are consistent with ACC. Our proposed ACC significantly outperforms its counterpart ABC, with a detailed experimental analysis in Section~\ref{sec:analysis}.

\begin{table*}[t]
\centering
\scriptsize
\caption{\textbf{mIoU results for 1-shot and 5-shot scenarios for both traditional FSS and CD-FSS methods across the four CD-FSS benchmarks.} Bold values indicate the best performance among all methods. ``\dag" means the results are reproduced by ourselves.}
\label{tab:comparison-experiments}
\begin{tabular}{cc|cc|cc|cc|cc|cc}
\toprule
{\multirow{2}{*}{Methods}} & {\multirow{2}{*}{Publication}} &
\multicolumn{2}{c}{Deepglobe} &
  \multicolumn{2}{c}{ISIC} &
  \multicolumn{2}{c}{Chest X-Ray} &
  \multicolumn{2}{c}{FSS-1000} &
  \multicolumn{2}{c}{Average} \\ \cmidrule{3-12} 
 & & 1-shot & 5-shot & 1-shot & 5-shot & 1-shot & 5-shot & 1-shot & 5-shot & 1-shot & 5-shot \\ \midrule
 \multicolumn{12}{c}{Few-shot Semantic Segmentation Methods} \\ \midrule
PGNet~\cite{zhang2019pyramid} & ICCV 2019 & 10.73 & 12.36 & 21.86 & 21.25 & 33.95 & 27.96 & 62.42 & 62.74 & 32.24 & 31.08 \\
PANet~\cite{wang2019panet} & ICCV 2019 & 36.55 & 45.43 & 25.29 & 33.99 & 57.75 & 69.31 & 69.15 & 71.68 & 47.19 & 55.10 \\
CaNet\cite{zhang2019canet} & CVPR 2019 & 22.32 & 23.07 & 25.16 & 28.22 & 28.35 & 28.62 & 70.67 & 72.03 & 36.63 & 37.99 \\
RPMMs~\cite{yang2020prototype} & ECCV 2020  & 12.99 & 13.47 & 18.02 & 20.04 & 30.11 & 30.82 & 65.12 & 67.06 & 31.56 & 32.85 \\
PFENet~\cite{tian2020prior} & TPAMI 2020 & 16.88 & 18.01 & 23.50 & 23.83 & 27.22 & 27.57 & 70.87 & 70.52 & 34.62 & 34.98 \\
RePRI~\cite{boudiaf2021few} & CVPR 2021  & 25.03 & 27.41 & 23.27 & 26.23 & 65.08 & 65.48 & 70.96 & 74.23 & 46.09 & 48.34 \\
HSNet~\cite{min2021hypercorrelation} & ICCV 2021  & 29.65 & 35.08 & 31.20 & 35.10 & 51.88 & 54.36 & 77.53 & 80.99 & 47.57 & 51.38 \\
SSP~\cite{fan2022self} & ECCV 2022 & 40.48 & 49.66 & 35.09 & 44.96 & 74.23 & 80.51 & 79.03 & 80.56 & 57.20 & 63.92 \\ \midrule
 \multicolumn{12}{c}{Cross-domain Few-shot Semantic Segmentation Methods} \\ \midrule
PATNet~\cite{lei2022cross} & ECCV 2022   & 37.89  & 42.97  & 41.16  & 53.58  & 66.61  & 70.20  & 78.59  & 81.23  & 56.06  & 61.99  \\
IFA\dag~\cite{nie2024cross}  & CVPR 2024 & 37.73  & 44.83  & 44.55  & 49.20  & 80.03  & 82.77  & 79.97  & 82.95  & 60.57  & 64.94  \\
DR-Adaptor~\cite{su2024domain} & CVPR 2024 & 41.29  & \textbf{50.12}  & 40.77  & 48.87  & 82.35  & 82.31  & 79.05  & 80.40  & 60.86  & 65.42  \\
ABCDFSS~\cite{herzog2024adapt} & CVPR 2024 & \textbf{42.60}  & 49.00  & 45.70  & 53.30  & 79.80  & 81.40  & 74.60  & 76.20  & 60.70  & 65.00  \\
PMNNet~\cite{chen2024pixel} & WACV 2024 & 37.10  & 41.60  & \textbf{51.20}  & 54.50  & 70.40  & 74.00  & \textbf{84.60}  & \textbf{86.30}  & 60.83  & 64.10  \\
\midrule
 \multicolumn{12}{c}{SAM-based Reference Segmentation Methods} \\ \midrule
VRP-SAM\dag~\cite{sun2024vrp} & CVPR 2024  & 30.62  & 33.43 & 40.81  & 42.47  & 77.33  & 80.70  & 81.25  & 82.71  & 57.50  & 59.83 \\
APSeg~\cite{he2024apseg} & CVPR 2024 & 35.94  & 39.98  & 45.43  & 53.98  & 84.10  & 84.50  & 79.71  & 81.90  & 61.30  & 65.09  \\
\midrule
 \multicolumn{12}{c}{Our Correspondence as Video Methods} \\ \midrule
CAV-SAM \bblue{(Ours)} & -    & 39.11  & 44.16  & 50.36  & \textbf{59.11}  & \textbf{86.97}  & \textbf{88.91}  & 79.78  & 84.38  & \textbf{64.06}  & \textbf{69.14}  \\ \bottomrule
\end{tabular}
\end{table*}

\noindent\textbf{Geometric Alignment.}
After the test-time lightweight fine-tuning on SAM2's image encoder as shown in Fig.~\ref{fig:TTGA-fine-tune}, the prototype vector $\boldsymbol p_r$ are employed to activate the pseudo video frames $I_v^1,I_v^2,\dots,I_v^{N_v}$ to obtain pseudo-labels $\hat M_v^1,\hat M_v^2,\dots,\hat M_v^{N_v}$, which will serve as additional prompts for SAM2, aligning with the geometric variations.
Specifically, we prompt the first half of the frames.

\subsection{Discussions}

\textbf{Complexity.}
Our approach establishes a universal framework for reference segmentation that can be implemented with any iVOS and diffusion model.
Specifically, SAM2 achieves real-time performance on modern GPUs~\cite{ravi2024sam}, while advancements in diffusion models are progressively decreasing computational cost~\cite{xue2024accelerating,huang2024tfmq}. The complexity can be tailored based on the backbone selection, enabling flexible trade-offs between performance and efficiency.

\noindent\textbf{Semantic Consistency.}
While all \bblue{few-shot} segmentation datasets~\cite{shaban2017one,nguyen2019feature,lei2022cross} presuppose the presence of identical class in both reference and target images, real-world \bblue{reference segmentation data} often violate this assumption. As depicted in Fig.~\ref{fig:semantic_consistency}, our method exhibits resilience when the target image lacks the reference class, which is attributed to two key components: The DBST module, by utilizing the text embeddings of Stable Diffusion, ensures semantic coherence. When confronted with semantic discrepancies, it produces a meaningless sequence, thereby preventing the iVOS model from initiating incorrect segmentations. Furthermore, the TTGA module with prototype vectors encapsulating class semantics, remains inactive when presented with semantically inconsistent input, thus avoiding the generation of misleading prompts for the iVOS model.


%% file: sec/4_experiments.tex
\section{Experiments}

\subsection{Datasets}

We evaluate our methods following the approach in~\cite{lei2022cross}. This benchmark comprises images and pixel-level annotations from various datasets: FSS-1000~\cite{li2020fss}, DeepGlobe~\cite{demir2018deepglobe}, ISIC2018~\cite{codella2019skin,tschandl2018ham10000}, and Chest X-ray~\cite{candemir2013lung,jaeger2013automatic}. Spanning both natural and medical image domains, these datasets provide substantial domain diversity for extensive evaluation.

\begin{figure*}[t]
    \centering
    \includegraphics[width=\linewidth]{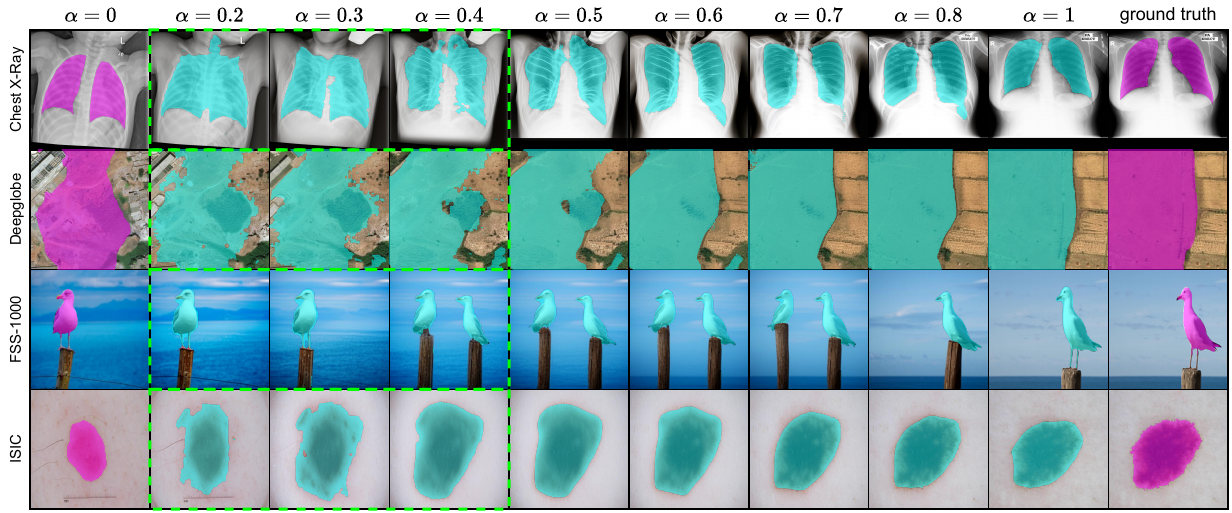}
    \caption{\textbf{Visualization of 1-shot segmentation results across four datasets.} The \textcolor[RGB]{200,0,200}{\textbf{magenta}} mask denotes ground truth for reference-target images, while the \textcolor[RGB]{0,200,200}{\textbf{cyan}} mask shows predictions. \textcolor[RGB]{0,200,0}{\textbf{Green}} dashed lines outline frames where the TTGA module prompts SAM2.}
    \label{fig:vis-all}
\end{figure*}


\subsection{Implementation Details}

We conducted our evaluation using the \verb|tiny| version of SAM2 . For the DBST module, we employed a learning rate of \(2 \times 10^{-4}\) during the training of LoRA and completed 200 steps with a rank of 16. During the generation process, we executed 20 steps of DDIM inversion for each frame. The semantic transition sequence length was set to \(N_v = 9\), with \(\alpha\) values ranging uniformly from 0.2 to 0.8 excluding the endpoints 0 and 1. For the TTGA module, we trained for 100 steps per reference image with an initial learning rate of \(1 \times 10^{-3}\) with cosine annealing for the scheduling.

\subsection{Comparison Experiments}

We compared our CAV-SAM with several reference segmentation methods based on meta-learning, including FSS, CD-FSS, and SAM-based reference segmentation approaches.
Tab.~\ref{tab:comparison-experiments} provides a detailed comparison of our method with these approaches.
Our method significantly enhances performance by approximately $5\%$ over SOTA techniques without the need for extensive meta-training. This improvement is particularly pronounced with the most substantial gains observed on the Chest X-Ray dataset.
However, our method does not achieve optimal results on the Deepglobe dataset. We hypothesize that this is due to the fact that SAM is designed for object segmentation, whereas Deepglobe focuses on segmenting regions.

\subsection{Ablation Studies}
\label{sec:ablation}
\begin{table}[t]
    \centering
    \footnotesize
    \caption{\textbf{Ablation study on key components of CAV-SAM}. Results are averaged over 4 datasets for 1-shot.}
    \label{tab:ablation-studies}
    \begin{tabular}{c|cc|c}
    \toprule
       iVOS Model & DBST & TTGA & mIoU \\ \midrule
       {\multirow{3}{*}{SAM2~\cite{ravi2024sam}}} &  - & - & 60.68 \\
        & \checkmark & - & 62.68 \\
       &  \checkmark & \checkmark & 64.06 \\ \midrule
       {\multirow{3}{*}{DEVA~\cite{cheng2023tracking}}} &  - & - & 49.57 \\
        & \checkmark & - & 57.31 \\
       &  \checkmark & \checkmark & 59.93 \\ \bottomrule
    \end{tabular}
\end{table}

We conducted ablation studies on three key components: the iVOS model selection, the DBST module, and the TTGA module.
For experiments without DBST, we utilized the concatenation baseline as shown in Fig.~\ref{fig:challenges}. 
For the experiments without TTGA, we limited our prompts to the first frame.
The results illustrated in Tab.~\ref{tab:ablation-studies} indicate that our strong baseline nearly matches the performance of SOTA CD-FSS methods, underscoring the effectiveness of embodying correspondence as video. Additionally, DBST and TTGA improved overall performance as well.

\subsection{More Analysis}
\label{sec:analysis}

\begin{table}[t]
    \centering
    \footnotesize
    \caption{\textbf{Evaluation of heuristic approaches to generate pseudo video frames.} Results are averaged over 4 datasets for 1-shot.}
    \label{tab:heuristic-approaches}
    \begin{tabular}{c|cccc}
    \toprule
     & Concat & Mixup & Affine & CAV-SAM  \\ \midrule
   mIoU &  60.68 & 52.21 & 56.84 & 64.06 \\ \bottomrule 
    \end{tabular}
\end{table}

\textbf{Heuristic Approaches.}
The key to treating correspondence between images as video is how to transition discrete images into a smooth transition sequence.
In this section, we investigate whether heuristic approaches depicted in Fig.~\ref{fig:challenges} can effectively address this challenge. 
To mitigate semantic discrepancy, we employ mixup~\cite{zhang2017mixup} to create sequences blending the two images at varying levels. 
To enhance the iVOS model's ability to capture geometric variance, we use affine augmentation to create sequences with geometric diversity. Results presented in Tab.~\ref{tab:heuristic-approaches} reveal that although these heuristic methods have theoretical potential, they even underperform the concatenation baseline. We attribute this to their inability to  simulate natural videos.

\begin{table}[t]
    \centering
    \footnotesize
    \caption{\textbf{Evaluation of ACC and ABC for the test-time fine-tuning approach.} Results are averaged over 4 datasets for 1-shot.}
    \label{tab:acc-abc}
    \begin{tabular}{c|ccc}
    \toprule
     & DBST & +TTGA(ACC) & +TTGA(ABC)  \\ \midrule
   mIoU & 62.08 &  64.06 & 61.54 \\ \bottomrule 
    \end{tabular}
\end{table}

\noindent\textbf{Effectiveness of Augmentative Cyclic Consistency.}
In test-time fine-tuning, we propose augmentative cyclical consistency (ACC). 
This technique generates an augmented image $I_r^{\text{aug}}$ and utilizes the prototype vector $\boldsymbol p_r$ to produce an augmentative pseudo-label $\hat M_r^{\text{aug}}$. $\hat M_r^{\text{aug}}$ is then used to derive an augmented prototype vector $\boldsymbol p_r^{\text{aug}}$. 

We also explore an alternative method, augmentative bi-directional consistency (ABC). As shown in Fig.~\ref{fig:acc-abc}, while ABC employs the augmented ground truth $M_r^{\text{aug}}$ to derive $\boldsymbol p_r^{\text{aug}}$, all other processes are consistent with those in ACC. 
Experimental results presented in Tab.~\ref{tab:acc-abc} indicate that ACC outperforms ABC. 
This improvement can be attributed to the fact that cyclical consistency poses a more challenging task than bi-directional consistency, thus improving the robustness of the prototype vectors refined by TTGA.

\noindent\textbf{Qualitative Results.}
In Fig.~\ref{fig:vis-all} we present visualizations of 1-shot segmentation results for all four datasets, alongside the intermediate segmentation results for each frame in the pseudo video generated by the DBST module. The \textcolor[RGB]{200,0,200}{\textbf{magenta}} mask indicates the ground truth for the reference-target images, while the \textcolor[RGB]{0,200,200}{\textbf{cyan}} mask represents the predicted segmentations. Regions outlined by \textcolor[RGB]{0,200,0}{\textbf{green}} dashed lines highlight the frames where the TTGA module employs additional prompts on SAM2 for geometric alignment. 

Our DBST module effectively generates natural pseudo video sequences, while the TTGA module effectively refines prototype vectors using only one labeled reference image, obtaining high-quality pseudo-label as prompts to facilitate geometric alignment of the iVOS model.

\begin{figure}[t]
    \centering
    \includegraphics[width=\linewidth]{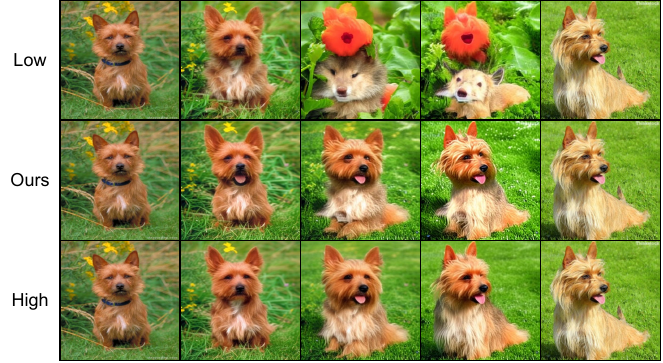}
    \caption{\textbf{Different semantic transformation sequences produced by different parameter and module configurations.} As the level of semantic transformation increases, the outputs become more aligned with human visual perception, though this comes at a significantly greater time cost.}
    \label{fig:parameter-sensitivity-example}
\end{figure}

\begin{figure}[t]
    \centering
    \includegraphics[width=\linewidth]{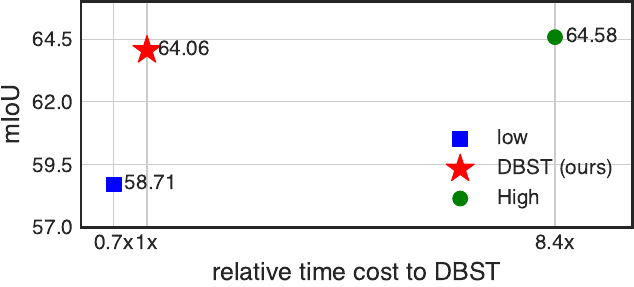}
    \caption{\textbf{Evaluation of different parameter configurations to generated semantic transformation sequences.} Our DBST module effectively achieves an optimal balance between time expenditure and overall segmentation performance.}
    \label{fig:parameter-sensitivity-result}
\end{figure}

\subsection{Parameter Sensitivity}
\label{sec:parameter-sensitivity}
The parameter sensitivity of our approach is predominantly influenced by the DBST module because of the introduction of the diffusion model.
Diffusion models are notably sensitive to parameter adjustments. Different parameters can lead to markedly distinct outcomes, and even identical parameter sets may yield variable results across different images. 
We will demonstrate that our approach is relatively robust to the quality of the pseudo video. 
Furthermore, the DBST module is capable of achieving a trade-off between segmentation performance and computational overhead.

The primary goal of most diffusion-based methods like DiffMorpher~\cite{zhang2024diffmorpher} is to generate intermediate images that are perceptually meaningful to human, necessitating various post-processing modules. In contrast, our main objective is simply to generate a natural pseudo video sequence, which only requires basic low-level semantic transformations. By removing non-essential enhancements and optimizing parameters for LoRA training and DDIM inversion, we substantially reduce the inference costs of the DBST module.

As illustrated in Fig.~\ref{fig:parameter-sensitivity-example}, we designed three distinct sets of parameter and post-processing module configuration,
representing varying levels of semantic transformation. The experimental results in Fig.~\ref{fig:parameter-sensitivity-result} validate that as long as the pseudo video sequence maintains acceptable, our DBST module remains robust.
And our DBST module achieves optimal performance and cost balance by maximizing time efficiency while maintaining high segmentation accuracy.

%% file: sec/5_conclusion.tex
\section{Conclusion}

Our proposed method, termed CAV-SAM, enhances SAM2 by embodying the correspondence between reference-target image pairs into a pseudo video sequence, overcoming the limitations of traditional reference segmentation approaches reliant on extensive meta-learning. By introducing two key modules---DBST for generating smooth semantic transitions using the diffusion model and TTGA for aligning geometric variations by test-time lightweight fine-tuning to perform additional prompts on SAM2---we achieve significant performance improvements. Our evaluations demonstrate that CAV-SAM surpasses state-of-the-art methods in CD-FSS benchmarks, showcasing its robustness and effectiveness in real-world applications. Our proposed method not only streamlines the adaptation of SAM2 for downstream tasks in the wild, but also contributes possible insights for future interactive video object segmentation research.

%% file: main.bbl
\begin{thebibliography}{50}
\providecommand{\natexlab}[1]{#1}
\providecommand{\url}[1]{\texttt{#1}}
\expandafter\ifx\csname urlstyle\endcsname\relax
  \providecommand{\doi}[1]{doi: #1}\else
  \providecommand{\doi}{doi: \begingroup \urlstyle{rm}\Url}\fi

\bibitem[Boudiaf et~al.(2021)Boudiaf, Kervadec, Masud, Piantanida, Ben~Ayed, and Dolz]{boudiaf2021few}
Malik Boudiaf, Hoel Kervadec, Ziko~Imtiaz Masud, Pablo Piantanida, Ismail Ben~Ayed, and Jose Dolz.
\newblock Few-shot segmentation without meta-learning: A good transductive inference is all you need?
\newblock In \emph{Proceedings of the IEEE/CVF Conference on Computer Vision and Pattern Recognition}, pages 13979--13988, 2021.

\bibitem[Candemir et~al.(2013)Candemir, Jaeger, Palaniappan, Musco, Singh, Xue, Karargyris, Antani, Thoma, and McDonald]{candemir2013lung}
Sema Candemir, Stefan Jaeger, Kannappan Palaniappan, Jonathan~P Musco, Rahul~K Singh, Zhiyun Xue, Alexandros Karargyris, Sameer Antani, George Thoma, and Clement~J McDonald.
\newblock Lung segmentation in chest radiographs using anatomical atlases with nonrigid registration.
\newblock \emph{IEEE Transactions on Medical Imaging}, 33\penalty0 (2):\penalty0 577--590, 2013.

\bibitem[Chen et~al.(2024)Chen, Dong, Lu, Yu, and Han]{chen2024pixel}
Hao Chen, Yonghan Dong, Zheming Lu, Yunlong Yu, and Jungong Han.
\newblock Pixel matching network for cross-domain few-shot segmentation.
\newblock In \emph{Proceedings of the IEEE/CVF Winter Conference on Applications of Computer Vision}, pages 978--987, 2024.

\bibitem[Cheng et~al.(2022)Cheng, Lang, and Han]{cheng2022holistic}
Gong Cheng, Chunbo Lang, and Junwei Han.
\newblock Holistic prototype activation for few-shot segmentation.
\newblock \emph{IEEE Transactions on Pattern Analysis and Machine Intelligence}, 45\penalty0 (4):\penalty0 4650--4666, 2022.

\bibitem[Cheng et~al.(2023{\natexlab{a}})Cheng, Oh, Price, Schwing, and Lee]{cheng2023tracking}
Ho~Kei Cheng, Seoung~Wug Oh, Brian Price, Alexander Schwing, and Joon-Young Lee.
\newblock Tracking anything with decoupled video segmentation.
\newblock In \emph{Proceedings of the IEEE/CVF International Conference on Computer Vision}, pages 1316--1326, 2023{\natexlab{a}}.

\bibitem[Cheng et~al.(2023{\natexlab{b}})Cheng, Li, Xu, Li, Yang, Wang, and Yang]{cheng2023segment}
Yangming Cheng, Liulei Li, Yuanyou Xu, Xiaodi Li, Zongxin Yang, Wenguan Wang, and Yi Yang.
\newblock Segment and track anything.
\newblock \emph{arXiv preprint arXiv:2305.06558}, 2023{\natexlab{b}}.

\bibitem[Codella et~al.(2019)Codella, Rotemberg, Tschandl, Celebi, Dusza, Gutman, Helba, Kalloo, Liopyris, Marchetti, et~al.]{codella2019skin}
Noel Codella, Veronica Rotemberg, Philipp Tschandl, M~Emre Celebi, Stephen Dusza, David Gutman, Brian Helba, Aadi Kalloo, Konstantinos Liopyris, Michael Marchetti, et~al.
\newblock Skin lesion analysis toward melanoma detection 2018: A challenge hosted by the international skin imaging collaboration (isic).
\newblock \emph{arXiv preprint arXiv:1902.03368}, 2019.

\bibitem[Demir et~al.(2018)Demir, Koperski, Lindenbaum, Pang, Huang, Basu, Hughes, Tuia, and Raskar]{demir2018deepglobe}
Ilke Demir, Krzysztof Koperski, David Lindenbaum, Guan Pang, Jing Huang, Saikat Basu, Forest Hughes, Devis Tuia, and Ramesh Raskar.
\newblock Deepglobe 2018: A challenge to parse the earth through satellite images.
\newblock In \emph{Proceedings of the IEEE Conference on Computer Vision and Pattern Recognition workshops}, pages 172--181, 2018.

\bibitem[Fan et~al.(2022)Fan, Pei, Tai, and Tang]{fan2022self}
Qi Fan, Wenjie Pei, Yu-Wing Tai, and Chi-Keung Tang.
\newblock Self-support few-shot semantic segmentation.
\newblock In \emph{European Conference on Computer Vision}, pages 701--719. Springer, 2022.

\bibitem[He et~al.(2024)He, Zhang, Zhuo, Shen, Yang, Deng, and Sun]{he2024apseg}
Weizhao He, Yang Zhang, Wei Zhuo, Linlin Shen, Jiaqi Yang, Songhe Deng, and Liang Sun.
\newblock Apseg: Auto-prompt network for cross-domain few-shot semantic segmentation.
\newblock In \emph{Proceedings of the IEEE/CVF Conference on Computer Vision and Pattern Recognition}, pages 23762--23772, 2024.

\bibitem[Herzog(2024)]{herzog2024adapt}
Jonas Herzog.
\newblock Adapt before comparison: A new perspective on cross-domain few-shot segmentation.
\newblock In \emph{Proceedings of the IEEE/CVF Conference on Computer Vision and Pattern Recognition}, pages 23605--23615, 2024.

\bibitem[Ho et~al.(2020)Ho, Jain, and Abbeel]{ho2020denoising}
Jonathan Ho, Ajay Jain, and Pieter Abbeel.
\newblock Denoising diffusion probabilistic models.
\newblock \emph{Advances in Neural Information Processing Systems}, 33:\penalty0 6840--6851, 2020.

\bibitem[Hu et~al.(2021)Hu, Shen, Wallis, Allen-Zhu, Li, Wang, Wang, and Chen]{hu2021lora}
Edward~J Hu, Yelong Shen, Phillip Wallis, Zeyuan Allen-Zhu, Yuanzhi Li, Shean Wang, Lu Wang, and Weizhu Chen.
\newblock Lora: Low-rank adaptation of large language models.
\newblock \emph{arXiv preprint arXiv:2106.09685}, 2021.

\bibitem[Huang et~al.(2024)Huang, Gong, Liu, Chen, and Liu]{huang2024tfmq}
Yushi Huang, Ruihao Gong, Jing Liu, Tianlong Chen, and Xianglong Liu.
\newblock Tfmq-dm: Temporal feature maintenance quantization for diffusion models.
\newblock In \emph{Proceedings of the IEEE/CVF Conference on Computer Vision and Pattern Recognition}, pages 7362--7371, 2024.

\bibitem[Jaeger et~al.(2013)Jaeger, Karargyris, Candemir, Folio, Siegelman, Callaghan, Xue, Palaniappan, Singh, Antani, et~al.]{jaeger2013automatic}
Stefan Jaeger, Alexandros Karargyris, Sema Candemir, Les Folio, Jenifer Siegelman, Fiona Callaghan, Zhiyun Xue, Kannappan Palaniappan, Rahul~K Singh, Sameer Antani, et~al.
\newblock Automatic tuberculosis screening using chest radiographs.
\newblock \emph{IEEE Transactions on Medical Imaging}, 33\penalty0 (2):\penalty0 233--245, 2013.

\bibitem[Kirillov et~al.(2023)Kirillov, Mintun, Ravi, Mao, Rolland, Gustafson, Xiao, Whitehead, Berg, Lo, et~al.]{kirillov2023segment}
Alexander Kirillov, Eric Mintun, Nikhila Ravi, Hanzi Mao, Chloe Rolland, Laura Gustafson, Tete Xiao, Spencer Whitehead, Alexander~C Berg, Wan-Yen Lo, et~al.
\newblock Segment anything.
\newblock In \emph{Proceedings of the IEEE/CVF International Conference on Computer Vision}, pages 4015--4026, 2023.

\bibitem[Lang et~al.(2022)Lang, Cheng, Tu, and Han]{lang2022learning}
Chunbo Lang, Gong Cheng, Binfei Tu, and Junwei Han.
\newblock Learning what not to segment: A new perspective on few-shot segmentation.
\newblock In \emph{Proceedings of the IEEE/CVF Conference on Computer Vision and Pattern Recognition}, pages 8057--8067, 2022.

\bibitem[Lang et~al.(2023)Lang, Cheng, Tu, Li, and Han]{lang2023base}
Chunbo Lang, Gong Cheng, Binfei Tu, Chao Li, and Junwei Han.
\newblock Base and meta: A new perspective on few-shot segmentation.
\newblock \emph{IEEE Transactions on Pattern Analysis and Machine Intelligence}, 45\penalty0 (9):\penalty0 10669--10686, 2023.

\bibitem[Lei et~al.(2022)Lei, Zhang, He, Chen, Du, and Lu]{lei2022cross}
Shuo Lei, Xuchao Zhang, Jianfeng He, Fanglan Chen, Bowen Du, and Chang-Tien Lu.
\newblock Cross-domain few-shot semantic segmentation.
\newblock In \emph{European Conference on Computer Vision}, pages 73--90. Springer, 2022.

\bibitem[Li et~al.(2021)Li, Jampani, Sevilla-Lara, Sun, Kim, and Kim]{li2021adaptive}
Gen Li, Varun Jampani, Laura Sevilla-Lara, Deqing Sun, Jonghyun Kim, and Joongkyu Kim.
\newblock Adaptive prototype learning and allocation for few-shot segmentation.
\newblock In \emph{Proceedings of the IEEE/CVF Conference on Computer Vision and Pattern Recognition}, pages 8334--8343, 2021.

\bibitem[Li et~al.(2020)Li, Wei, Chen, Tai, and Tang]{li2020fss}
Xiang Li, Tianhan Wei, Yau~Pun Chen, Yu-Wing Tai, and Chi-Keung Tang.
\newblock Fss-1000: A 1000-class dataset for few-shot segmentation.
\newblock In \emph{Proceedings of the IEEE/CVF Conference on Computer Vision and Pattern Recognition}, pages 2869--2878, 2020.

\bibitem[Liu et~al.(2022)Liu, Liu, Yao, and Han]{liu2022intermediate}
Yuanwei Liu, Nian Liu, Xiwen Yao, and Junwei Han.
\newblock Intermediate prototype mining transformer for few-shot semantic segmentation.
\newblock \emph{Advances in Neural Information Processing Systems}, 35:\penalty0 38020--38031, 2022.

\bibitem[Lu et~al.(2021)Lu, He, Zhu, Zhang, Song, and Xiang]{lu2021simpler}
Zhihe Lu, Sen He, Xiatian Zhu, Li Zhang, Yi-Zhe Song, and Tao Xiang.
\newblock Simpler is better: Few-shot semantic segmentation with classifier weight transformer.
\newblock In \emph{Proceedings of the IEEE/CVF International Conference on Computer Vision}, pages 8741--8750, 2021.

\bibitem[Mazurowski et~al.(2023)Mazurowski, Dong, Gu, Yang, Konz, and Zhang]{mazurowski2023segment}
Maciej~A Mazurowski, Haoyu Dong, Hanxue Gu, Jichen Yang, Nicholas Konz, and Yixin Zhang.
\newblock Segment anything model for medical image analysis: an experimental study.
\newblock \emph{Medical Image Analysis}, 89:\penalty0 102918, 2023.

\bibitem[Min et~al.(2021)Min, Kang, and Cho]{min2021hypercorrelation}
Juhong Min, Dahyun Kang, and Minsu Cho.
\newblock Hypercorrelation squeeze for few-shot segmentation.
\newblock In \emph{Proceedings of the IEEE/CVF International Conference on Computer Vision}, pages 6941--6952, 2021.

\bibitem[Nguyen and Todorovic(2019)]{nguyen2019feature}
Khoi Nguyen and Sinisa Todorovic.
\newblock Feature weighting and boosting for few-shot segmentation.
\newblock In \emph{Proceedings of the IEEE/CVF international conference on computer vision}, pages 622--631, 2019.

\bibitem[Nie et~al.(2024)Nie, Xing, Zhang, Yan, Xiao, Tan, Kot, and Lu]{nie2024cross}
Jiahao Nie, Yun Xing, Gongjie Zhang, Pei Yan, Aoran Xiao, Yap-Peng Tan, Alex~C Kot, and Shijian Lu.
\newblock Cross-domain few-shot segmentation via iterative support-query correspondence mining.
\newblock In \emph{Proceedings of the IEEE/CVF Conference on Computer Vision and Pattern Recognition}, pages 3380--3390, 2024.

\bibitem[Otsu et~al.(1975)]{otsu1975threshold}
Nobuyuki Otsu et~al.
\newblock A threshold selection method from gray-level histograms.
\newblock \emph{Automatica}, 11\penalty0 (285-296):\penalty0 23--27, 1975.

\bibitem[Peng et~al.(2023)Peng, Tian, Wu, Wang, Liu, Su, and Jia]{peng2023hierarchical}
Bohao Peng, Zhuotao Tian, Xiaoyang Wu, Chengyao Wang, Shu Liu, Jingyong Su, and Jiaya Jia.
\newblock Hierarchical dense correlation distillation for few-shot segmentation.
\newblock In \emph{Proceedings of the IEEE/CVF Conference on Computer Vision and Pattern Recognition}, pages 23641--23651, 2023.

\bibitem[Raji{\v{c}} et~al.(2023)Raji{\v{c}}, Ke, Tai, Tang, Danelljan, and Yu]{rajivc2023segment}
Frano Raji{\v{c}}, Lei Ke, Yu-Wing Tai, Chi-Keung Tang, Martin Danelljan, and Fisher Yu.
\newblock Segment anything meets point tracking.
\newblock \emph{arXiv preprint arXiv:2307.01197}, 2023.

\bibitem[Ravi et~al.(2024)Ravi, Gabeur, Hu, Hu, Ryali, Ma, Khedr, R{\"a}dle, Rolland, Gustafson, et~al.]{ravi2024sam}
Nikhila Ravi, Valentin Gabeur, Yuan-Ting Hu, Ronghang Hu, Chaitanya Ryali, Tengyu Ma, Haitham Khedr, Roman R{\"a}dle, Chloe Rolland, Laura Gustafson, et~al.
\newblock Sam 2: Segment anything in images and videos.
\newblock \emph{arXiv preprint arXiv:2408.00714}, 2024.

\bibitem[Shaban et~al.(2017)Shaban, Bansal, Liu, Essa, and Boots]{shaban2017one}
Amirreza Shaban, Shray Bansal, Zhen Liu, Irfan Essa, and Byron Boots.
\newblock One-shot learning for semantic segmentation.
\newblock \emph{arXiv preprint arXiv:1709.03410}, 2017.

\bibitem[Song et~al.(2020{\natexlab{a}})Song, Meng, and Ermon]{song2020denoising}
Jiaming Song, Chenlin Meng, and Stefano Ermon.
\newblock Denoising diffusion implicit models.
\newblock \emph{arXiv preprint arXiv:2010.02502}, 2020{\natexlab{a}}.

\bibitem[Song et~al.(2020{\natexlab{b}})Song, Sohl-Dickstein, Kingma, Kumar, Ermon, and Poole]{song2020score}
Yang Song, Jascha Sohl-Dickstein, Diederik~P Kingma, Abhishek Kumar, Stefano Ermon, and Ben Poole.
\newblock Score-based generative modeling through stochastic differential equations.
\newblock \emph{arXiv preprint arXiv:2011.13456}, 2020{\natexlab{b}}.

\bibitem[Su et~al.(2024)Su, Fan, Pei, Lu, and Chen]{su2024domain}
Jiapeng Su, Qi Fan, Wenjie Pei, Guangming Lu, and Fanglin Chen.
\newblock Domain-rectifying adapter for cross-domain few-shot segmentation.
\newblock In \emph{Proceedings of the IEEE/CVF Conference on Computer Vision and Pattern Recognition}, pages 24036--24045, 2024.

\bibitem[Sun et~al.(2024)Sun, Chen, Zhang, Zhang, Chen, Zhang, Ding, Wang, and Li]{sun2024vrp}
Yanpeng Sun, Jiahui Chen, Shan Zhang, Xinyu Zhang, Qiang Chen, Gang Zhang, Errui Ding, Jingdong Wang, and Zechao Li.
\newblock Vrp-sam: Sam with visual reference prompt.
\newblock In \emph{Proceedings of the IEEE/CVF Conference on Computer Vision and Pattern Recognition}, pages 23565--23574, 2024.

\bibitem[Tian et~al.(2020)Tian, Zhao, Shu, Yang, Li, and Jia]{tian2020prior}
Zhuotao Tian, Hengshuang Zhao, Michelle Shu, Zhicheng Yang, Ruiyu Li, and Jiaya Jia.
\newblock Prior guided feature enrichment network for few-shot segmentation.
\newblock \emph{IEEE Transactions on Pattern Analysis and Machine Intelligence}, 44\penalty0 (2):\penalty0 1050--1065, 2020.

\bibitem[Tschandl et~al.(2018)Tschandl, Rosendahl, and Kittler]{tschandl2018ham10000}
Philipp Tschandl, Cliff Rosendahl, and Harald Kittler.
\newblock The ham10000 dataset, a large collection of multi-source dermatoscopic images of common pigmented skin lesions.
\newblock \emph{Scientific data}, 5\penalty0 (1):\penalty0 1--9, 2018.

\bibitem[Wang et~al.(2019)Wang, Liew, Zou, Zhou, and Feng]{wang2019panet}
Kaixin Wang, Jun~Hao Liew, Yingtian Zou, Daquan Zhou, and Jiashi Feng.
\newblock Panet: Few-shot image semantic segmentation with prototype alignment.
\newblock In \emph{proceedings of the IEEE/CVF International Conference on Computer Vision}, pages 9197--9206, 2019.

\bibitem[Wang et~al.(2022)Wang, Duan, Wang, En, Fan, and Zhang]{wang2022remember}
Wenjian Wang, Lijuan Duan, Yuxi Wang, Qing En, Junsong Fan, and Zhaoxiang Zhang.
\newblock Remember the difference: Cross-domain few-shot semantic segmentation via meta-memory transfer.
\newblock In \emph{Proceedings of the IEEE/CVF Conference on Computer Vision and Pattern Recognition}, pages 7065--7074, 2022.

\bibitem[Wang et~al.(2023)Wang, Sun, and Zhang]{wang2023rethinking}
Yuan Wang, Rui Sun, and Tianzhu Zhang.
\newblock Rethinking the correlation in few-shot segmentation: A buoys view.
\newblock In \emph{Proceedings of the IEEE/CVF Conference on Computer Vision and Pattern Recognition}, pages 7183--7192, 2023.

\bibitem[Xue et~al.(2024)Xue, Liu, Chen, Zhang, Hu, Xie, and Li]{xue2024accelerating}
Shuchen Xue, Zhaoqiang Liu, Fei Chen, Shifeng Zhang, Tianyang Hu, Enze Xie, and Zhenguo Li.
\newblock Accelerating diffusion sampling with optimized time steps.
\newblock In \emph{Proceedings of the IEEE/CVF Conference on Computer Vision and Pattern Recognition}, pages 8292--8301, 2024.

\bibitem[Yang et~al.(2020)Yang, Liu, Li, Jiao, and Ye]{yang2020prototype}
Boyu Yang, Chang Liu, Bohao Li, Jianbin Jiao, and Qixiang Ye.
\newblock Prototype mixture models for few-shot semantic segmentation.
\newblock In \emph{Computer Vision--ECCV 2020: 16th European Conference, Glasgow, UK, August 23--28, 2020, Proceedings, Part VIII 16}, pages 763--778. Springer, 2020.

\bibitem[Yang et~al.(2023)Yang, Gao, Li, Gao, Wang, and Zheng]{yang2023track}
Jinyu Yang, Mingqi Gao, Zhe Li, Shang Gao, Fangjing Wang, and Feng Zheng.
\newblock Track anything: Segment anything meets videos.
\newblock \emph{arXiv preprint arXiv:2304.11968}, 2023.

\bibitem[Zhang et~al.(2019{\natexlab{a}})Zhang, Lin, Liu, Guo, Wu, and Yao]{zhang2019pyramid}
Chi Zhang, Guosheng Lin, Fayao Liu, Jiushuang Guo, Qingyao Wu, and Rui Yao.
\newblock Pyramid graph networks with connection attentions for region-based one-shot semantic segmentation.
\newblock In \emph{Proceedings of the IEEE/CVF International Conference on Computer Vision}, pages 9587--9595, 2019{\natexlab{a}}.

\bibitem[Zhang et~al.(2019{\natexlab{b}})Zhang, Lin, Liu, Yao, and Shen]{zhang2019canet}
Chi Zhang, Guosheng Lin, Fayao Liu, Rui Yao, and Chunhua Shen.
\newblock Canet: Class-agnostic segmentation networks with iterative refinement and attentive few-shot learning.
\newblock In \emph{Proceedings of the IEEE/CVF Conference on Computer Vision and Pattern Recognition}, pages 5217--5226, 2019{\natexlab{b}}.

\bibitem[Zhang et~al.(2023)Zhang, Liu, Cui, Huang, Lin, Yang, and Hu]{zhang2023comprehensive}
Chunhui Zhang, Li Liu, Yawen Cui, Guanjie Huang, Weilin Lin, Yiqian Yang, and Yuehong Hu.
\newblock A comprehensive survey on segment anything model for vision and beyond.
\newblock \emph{arXiv preprint arXiv:2305.08196}, 2023.

\bibitem[Zhang et~al.(2021)Zhang, Kang, Yang, and Wei]{zhang2021few}
Gengwei Zhang, Guoliang Kang, Yi Yang, and Yunchao Wei.
\newblock Few-shot segmentation via cycle-consistent transformer.
\newblock \emph{Advances in Neural Information Processing Systems}, 34:\penalty0 21984--21996, 2021.

\bibitem[Zhang(2017)]{zhang2017mixup}
Hongyi Zhang.
\newblock mixup: Beyond empirical risk minimization.
\newblock \emph{arXiv preprint arXiv:1710.09412}, 2017.

\bibitem[Zhang et~al.(2024)Zhang, Zhou, Xu, Dai, and Pan]{zhang2024diffmorpher}
Kaiwen Zhang, Yifan Zhou, Xudong Xu, Bo Dai, and Xingang Pan.
\newblock Diffmorpher: Unleashing the capability of diffusion models for image morphing.
\newblock In \emph{Proceedings of the IEEE/CVF Conference on Computer Vision and Pattern Recognition}, pages 7912--7921, 2024.

\end{thebibliography}
